\documentclass{article} 
\usepackage{iclr2020_conference,times}

\usepackage{amsmath,amsfonts,bm}

\def\eqref#1{equation~\ref{#1}}

\def\1{\bm{1}}

\DeclareMathAlphabet{\mathsfit}{\encodingdefault}{\sfdefault}{m}{sl}
\SetMathAlphabet{\mathsfit}{bold}{\encodingdefault}{\sfdefault}{bx}{n}

\usepackage{hyperref}
\usepackage{url}
\usepackage{graphicx}

\title{Teacher-Student Domain Adaptation for Biosensor Models}

\author{Lawrence G. Phillips, David B. Grimes, \& Yihan (Jessie) Li \thanks{Corresponding author} \\
Jawbone Health, London, UK\\
\texttt{\{lphillips, dgrimes, jessieli\}@jawbonehealth.com} \\
}

\iclrfinalcopy 
\begin{document}

\maketitle

\begin{abstract}
We present an approach to domain adaptation, addressing the case where data from the source domain is abundant, labelled data from the target domain is limited or non-existent, and a small amount of paired source-target data is available. The method is designed for developing deep learning models that detect the presence of medical conditions based on data from consumer-grade portable biosensors.
It addresses some of the key problems in this area, namely, the difficulty of acquiring large quantities of clinically labelled  data from the biosensor, and the noise and ambiguity that can affect the clinical labels. The idea is to pre-train an expressive model on a large dataset of labelled recordings from a sensor modality for which data is abundant, and then to adapt the model's lower layers so that its predictions on the target modality are similar to the original model's on paired examples from the source modality. We show that the pre-trained model's predictions provide a substantially better learning signal than the clinician-provided labels, and that this teacher-student technique significantly outperforms both a naive application of supervised deep learning and a label-supervised version of domain adaptation on a synthetic dataset and in a real-world case study on sleep apnea. By reducing the volume of data required and obviating the need for labels, our approach should reduce the cost associated with developing high-performance deep learning models for biosensors.
\end{abstract}

\section{Introduction}

We are interested in the following situation, frequently encountered when developing clinical machine learning models on biosensor data: We have abundant labelled data from one domain (the \emph{source} domain), little or no labelled data from another (the \emph{target} domain), and a small amount of paired data from the source and target domains. The prediction target is the same for both domains, and we want to train a model that works well in the target domain. This is a type of domain adaptation problem (there are other types: for instance, paired data might be unavailable, or data from the target domain might be abundant --- see \citep{DomainAdaptRev, SurvTranLer, AdaVizCats} for reviews).

The work of \citep{SupervisionTransfer} in computer vision, and of \citep{MicroSpeech1, MicroSpeech2} in speech recognition, tackles closely related but not identical problems. A core idea in all of these efforts, which is central to our approach as well, is teacher-student training. A ``teacher" model is trained in the source domain, and a ``student" model is trained to mimic its behaviour using paired source-target data. In \citep{SupervisionTransfer}, the student model learns to mimic the teacher's representations at some intermediate layer on a set of paired data: a head is then attached to the student and trained on labelled target domain data. In \citep{MicroSpeech1, MicroSpeech2} the approach is more direct: the student model is trained to reproduce the teacher's predictions on paired examples, where the target domain data is generated synthetically from source domain examples. All of this work assumes abundant paired data, and so does not attempt to control the sample complexity of student training. By contrast, we tackle the scenario where paired data is not available in large quantities: the only abundant data we assume is source domain data. We train the student to match the teacher's representations at some intermediate layer as well as to mimic its predictions, as with previous work. Additionally, we control student training sample complexity by initialising the student as a clone of the teacher, and by fine-tuning only a lower portion of its weights. This way, we recruit additional statistical power from the source domain.

Our method is likely to be especially useful on clinical time series data, for a number of reasons. Firstly, the technique obviates the need for large-scale acquisition of labelled clinical data, which can require the resources of a technology giant \citep{AppleAfib}. Secondly, clinical time series labels are often afflicted by noise and ambiguity. This can arise due to the nature of the target condition\footnote{For example, the exact start and end points of a sleep apnea event are not precisely defined}, or the limitations of the sensor used by the clinician for labelling, or human error. In many cases, then, clinical time series problems should be viewed as probabilistic classification problems, where the correct output is a distribution over labels rather than a hard label. One expects teacher-student training to work especially well on probabilistic classification problems: for each example $x_i$, the teacher can offer something close to the correct distribution $P(y|x_i)$ to the student, rather than a single sample from that distribution, as with standard supervised learning. This should greatly decrease variance, and as a consequence, decrease sample complexity \citep{DeepNetsAreNotDeep}.
Finally, our method is very straightforward compared with much of the work on domain adaptation, which tends to rely on adversarial training \citep{VarRecAdv, AdDiDoAd, CoDa}. We feel that this simplicity is an advantage in the medical domain. 

\subsection{Experiments and Findings}
We use our technique to develop a model for sleep apnea from a large dataset of labelled electrocardiogram (ECG) data, and a small amount of paired ECG/PPG data. For reproducibility, experiments on a synthetic dataset, where we perturb the ECG signal from a publicly available dataset in a way that loosely resembles the domain shift caused by using PPG, are also conducted.
 We compare against other approaches that might be taken given the same data (plus labels for the PPG), namely straightforward supervised learning, a naive application of the pre-trained model to a partly domain-invariant representation, and supervised domain adaptation (where the labels are used as targets rather than the pre-trained model's output). We find that teacher-student domain adaptation outperforms these alternatives both on the synthetic data and in the sleep apnea study, with pre-training and fine-tuning each contributing equally to the performance boost. A key finding is that, across all of our experiments, using the pre-trained model's predictions as targets significantly improves performance relative to using the clinical labels. We find that fine-tuning on the labels in the target domain (rather than the pre-trained model's output) actually degrades performance relative to naively applying the pre-trained model to the target domain. We expect that this can be explained largely by the variance-reduction argument given above. Other important (albeit less surprising) findings are that transfer learning gives a dramatic improvement over only using the target domain data, and that our particular domain adaptation architecture outperforms other flavours of transfer learning on the data we experiment with.

\section{Model}
\label{section:model}
Suppose we are given two datasets. One contains time series of features derived from some sensor (sensor A), together with time series of labels. The other contains time series of features derived from sensor A, and time series of the same features measured simultaneously by sensor B\footnote{In fact, the technique applies to any source of domain shift across sensors. For example, both sensors could be ECG, with one single-lead and the other multi-lead, or the sensors could come from different manufacturers.}. We denote a generic sensor A time series from this set $x^{(A)}$, and the same for sensor B. The second dataset may also contain additional features recorded by sensor B, along with those that it shares with sensor A: we denote these additional feature time series $x^{(B')}$. Our goal is to train a model that takes recordings from sensor B and returns a time series of predictions as to whether some physiological event is occurring or not --- a sequential binary classification problem.

We begin by pre-training a model on the first dataset. The model outputs a sequence of predictions $p$ given a sequence of input vectors $x$: $p=H(S(T(x)))$. $T$ (for ``tail") is a module that acts independently on each input vector $x_i$ in the sequence $x$ to produce a sequence of vectors $r$: $r_i = T(x_i)$. This sequence is passed to a sequence model $S$, which produces another sequence of vectors: $q = S(r)$. Finally, $H$ (for ``head") acts independently on each element of $q$, producing a sequence of predictions: $p_i = H(q_i)$.

\begin{figure}[t]
\centering
\includegraphics[width=14cm]{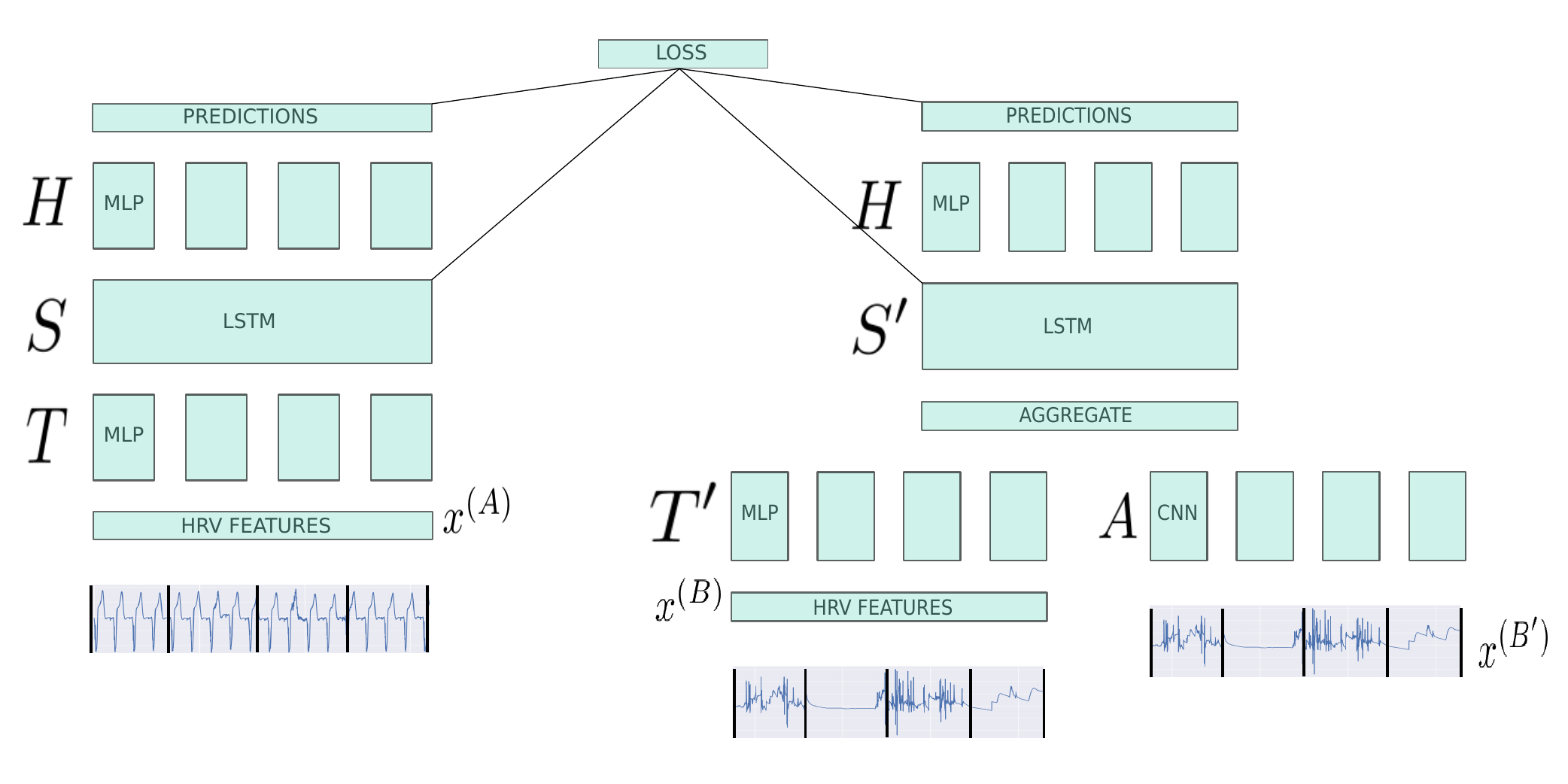}
\caption{Model architecture. The components are annotated according to the notation in the text (see Section \ref{section:model}).}
\label{fig:arch}
\end{figure}

After pre-training, we perform domain adaptation. $S$ and $T$ are cloned (we call the clones $S'$ and $T'$), and an auxiliary model $A$ that acts on $x^{(B')}$ is introduced. These parts come together to produce a sequence of predictions $p_B$ on data from sensor B as follows:
\begin{equation}
    p_B(x^{(B)}, x^{(B')}) = H\left(S'\left(\phi\left[T'(x^{(B)}), \,A(x^{(B')})\right]\right)\right),
\end{equation}
where $\phi$ is an aggregation function. The realisation of this architecture that we use for sleep apnea is pictured in Figure \ref{fig:arch}.

$T'$, $S'$, and $A$ are trained  on the activations and outputs of the pre-trained model: Given the paired inputs $x^{(A)}$, $x^{(B)}$, $x^{(B')}$, the loss is
\begin{equation}
    \label{loss}
    \mathcal{L} = \frac{1}{n}\sum_i[p_{Bi} - p_{Ai}] ^ 2 + \frac{\alpha}{nd}\sum_{ij}\left[S'\left(\phi\left[T'(x^{(B)}), \,A(x^{(B')})\right]\right)_{ij} - S(T(x^{(A)}))_{ij}\right]^2,
\end{equation}
where $p_{Ai} = H(S(T(x^{(A)}))_i)$, $\alpha$ is a real number, $n$ is the number of time steps in the example, $i$ indexes the time step, and $j$ indexes into the representation vector produced by $S$ or $S'$, which has dimension $d$. The $p_A$ and $p_B$ are logits. The loss function (\ref{loss}) is a hybrid between the losses from \citep{DeepNetsAreNotDeep} and \citep{SupervisionTransfer}, encouraging the domain adaptation model's internal representations of the target domain to be similar to the source domain, as well teaching the target model to mimic the source model's predictions. To control sample complexity, we do not update $H$ during domain adaptation: we are then free to make it highly expressive and have it absorb a large amount of information from the source domain.

\section{Case Study: Sleep Apnea}
In this section we demonstrate the real-world applicability of our method, using it to develop a model for sleep apnea. We collected recordings from wrist-worn PPG sensors, time-aligned with ECG, from subjects undergoing polysomnography (PSG) --- See Appendix \ref{section:apnea} for a description of sleep apnea, and of the study during which data collection was performed. We develop a model for detecting apnea events from PPG data, using this dataset for domain adaptation after pre-training an ECG model on the PhysioNet Computing in Cardiology Challenge 2018 (PCCC) dataset\footnote{\url{https://physionet.org/content/challenge-2018/1.0.0/}}, described in Appendix \ref{subsection:pccc-data}. The experiments reported here used data from 17 subjects. The PPG data was annotated by sleep experts using PSG data, and we use the labels to evaluate our label-free method and to compare it with label-supervised approaches.

To ensure that our work is reproducible, we also experiment on a synthetically altered version of PCCC, described in Appendix \ref{section:synth}, and to show which parts of our idea matter most, we perform an ablation study, found in Appendix \ref{section:abl}.
\subsection{Data}
As input to the model we use sequences of heart rate variability (HRV) features, a representation of ECG and PPG data that is used widely in machine learning, and in automated analysis generally \citep{PhilipsTL, Epilepsy, PSGPPG, Nature}.
 Although HRV features are partly domain-invariant, they are not completely so \citep{ThatFrom, EvalOfCoh}. In fact, domain shift may be most severe during the physiological events that we are interested in detecting. For instance, \citep{ApEvDiv} find that while under normal physiological conditions there is good agreement between ECG- and PPG-derived HRV features, the two diverge during apnea events. Moreover, signal from wearables may be especially prone to noise and distortion, further exacerbating domain shift.  We therefore expect a model trained on ECG-derived HRV features to perform suboptimally on PPG-derived HRV.
 HRV features are derived from sets of beat-to-beat intervals occurring within fixed time windows (referred to as \emph{epochs} in the sleep literature). Following standard practice, we divide the label time series into windows aligned with the HRV windows, and train our models to predict the label of each window (windows having less than a $75\%$ consensus as to the label are ignored).
 
\subsection{Model}
\label{subsection:model}
A high-level description of the model is given in Section \ref{section:model}. For $H$ we use a comparatively large multilayer perceptron (MLP), for $S$ we use a bi-directional long short-term memory module (LSTM) \citep{LSTM}, and for $T$ we use a smaller MLP. With these choices, the pre-trained model has the same structure as in \citep{PhilipsTL}, where conventional supervised transfer learning from ECG to PPG is studied. We calibrate the pre-trained model's predictions using Platt scaling \citep{PlattScaling} before using them to train the student --- this gives a small performance boost. $x^{(A)}$ and $x^{(B)}$ are sequences of HRV features derived from ECG and PPG respectively. $x^{(B')}_i$ contains the slice of raw PPG signal from which $x^{(B)}_i$ was computed. $A$ acts on the raw signal with a 1D convolutional neural network (CNN) \citep{CNN}, and then takes the mean along the time dimension. For the aggregation function $\phi$, we found simple addition to work well. We also experimented with an attention mechanism \citep{AIAYN}, but it performed poorly. In the loss function (\ref{loss}), $\alpha = 1$  was found to work well. For a schematic see Figure \ref{fig:arch}. We include the convolutional part because the raw signal may contain information about how the PPG-derived HRV features should be corrected, and there may be extra apnea-relevant information in the PPG signal beyond what is captured by HRV features.

\subsection{Results}
To show the effectiveness of teacher-student domain adaptation, and to gain some understanding of why it works well, we tested it alongside a number of other approaches. For all of these, we gathered predictions on all study subjects, using leave-one-out cross-validation at the subject level for approaches involving training on the PPG dataset, and computed metrics for all the predictions and labels taken together. An overview of the results is given in Table \ref{apnea-table}. The rows in the table correspond to the following experiments: For ``Label-Supervised (PPG only)", we trained a model solely on PPG data, with the clinical labels as targets, using straight-ahead supervised learning. The model architecture we used is identical to the teacher model in Figure \ref{fig:arch}\footnote{We also experimented with a convolutional LSTM acting on the raw PPG waveform along the lines of \citep{Jawbone}, and MLP-LSTM-MLP architecture acting on PPG peak-peak intervals.}. For ``Teacher-Student (PPG only)", we trained the same model on the same inputs as for ``Label-Supervised (PPG only)" but used the PCCC model's predictions as targets. For ``Label-Supervised Domain Adaptation", we took the approach described in Section \ref{subsection:model}, but with the cross entropy between predictions and clinical labels replacing the MSE between the two models' predictions\footnote{This is very close to the approach of \citep{SupervisionTransfer}, where shared representations are encouraged between teacher and student but labels are used to fit the target model}. For ``Label-Supervised Transfer Learning", we fine-tuned the entire PCCC model to adapt to PPG, using clinical labels as targets. For "Naive Transfer``, we directly applied the PCCC model to HRV features derived from PPG, with no fine-tuning. For ``Teacher-Student Transfer Learning" we unfroze $H$, and fine-tuned the entire PCCC model to adapt to PPG, using teacher-student training. For ``Teacher-Student Domain Adaptation", we did what was described in \ref{subsection:model}. Finally, ``ECG Input" is the teacher model applied to source-domain examples from the paired data. This represents the maximum theoretical performance of the student model. We see that Teacher-Student Domain Adaptation outperforms all other techniques, showing performance comparable to that achieved by applying the PCCC model to ECG data\footnote{Note that, although the experiments here are conducted using a small number of subjects, results are reported per-window and thus have low variance. For instance, using the formula from \citep{ROCVar} and the fact that the we have $\sim$6000 examples in the minority class, we can upper-bound the variance on the ROC AUC score for ``Teacher-Student Domain Adaptation" at $3.6\times 10 ^{-5}$.}.
\begin{table}[t]
\caption{Per-window validation metrics for sleep apnea.}
\label{apnea-table}
\begin{center}
\begin{tabular}{lll}
\multicolumn{1}{c}{\bf MODEL}  &\multicolumn{1}{c}{\bf ROC AUC}  &\multicolumn{1}{c}{\bf PR AUC}
\\ \hline \\
Label-Supervised (PPG only)  & 0.61 & 0.16 \\
Teacher-Student (PPG only) & 0.71 & 0.21 \\
Label-Supervised Domain Adaptation & 0.71 & 0.29 \\
Label-Supervised Transfer Learning & 0.71 & 0.30 \\
Naive Transfer & 0.75  & 0.34\\
Teacher-Student Transfer Learning & 0.78 & 0.42 \\
Teacher-Student Domain Adaptation & \bf{0.80}  & \bf{0.47}\\
(ECG input) & 0.83 & 0.47 \\
\end{tabular}
\end{center}
\end{table}

\subsubsection{Teacher-Student vs Label-Supervised}
One of our more interesting findings is that teacher-student approaches consistently outperform approaches that rely on the clinical labels, often to a striking degree. This is most starkly illustrated when we apply straightforward supervised learning to the PPG dataset, either training on the clinical labels or on the PCCC model's ECG-based predictions. Teacher-supervision improves performance in terms of area under the precision-recall curve by 30$\%$ in this case. We speculate that this partly due to hard labelling of ambiguous cases. Inter-rater disagreement is common in this domain, so the true distribution over labels given the input is not deterministic, and we can think of the labels seen during training as samples from the true distribution. This sampling will be a source of variance, which negatively affects generalisation. In this case, since naive transfer outperforms all label-supervised approaches, the negative effect of this variance outweighs the benefit from adjusting to domain shift during fine-tuning. Teacher-student domain adaptation works well partly because it does not incur this variance, as the model is presented with estimated probabilities rather than samples during training.

\section{Discussion}
We have shown that it is possible to recruit statistical power from large, publicly available datasets to produce models for sensors that do not have much readily available data associated with them, and for which labelled data may be completely unavailable. We found evidence that supervision from a pre-trained model may be strongly preferable to supervision from clinician-provided labels. We suspect that this finding may generalise to other forms of paired biosensor time series beyond PPG/ECG, and to medically relevant scenarios other than biosensor time series, and we look forward to future work exploring the potential and limits of teacher-student techniques in the clinical domain.

\subsubsection*{Acknowledgments}
The authors thank Dr Clete Anthony Kushida and the researchers at the Center for Sleep Medicine and the Center for Human Sleep Research, Stanford University, for drafting the original sleep protocol and conducting patient recruitment, Daniel Jones for the helpful comments and proof-reading the work, Dileep Goyal who led the hardware development effort of the wrist-worn device at Jawbone Health, Dr David Benaron and Kate Krasileva for helping with study design and coordination at Jawbone Health, and Hosain Rahman, CEO of Jawbone Health for his vision, leadership and support.

\bibliography{iclr2020_conference}

\begin{thebibliography}{30}
\providecommand{\natexlab}[1]{#1}
\providecommand{\url}[1]{\texttt{#1}}
\expandafter\ifx\csname urlstyle\endcsname\relax
  \providecommand{\doi}[1]{doi: #1}\else
  \providecommand{\doi}{doi: \begingroup \urlstyle{rm}\Url}\fi

\bibitem[Ba \& Caruana(2014)Ba and Caruana]{DeepNetsAreNotDeep}
J.~Ba and R.~Caruana.
\newblock Do deep nets really need to be deep?
\newblock In \emph{Advances in Neural Information Processing Systems 27}, pp.\
  2654--2662. Curran Associates, Inc., 2014.

\bibitem[Berry~et al(2017)]{AASM}
R.~B. Berry~et al.
\newblock Aasm scoring manual updates for 2017.
\newblock \emph{Journal of Clinical Medicine and Sleep}, 13:\penalty0
  1550--9389, 2017.

\bibitem[Brinbaum \& Klose(1957)Brinbaum and Klose]{ROCVar}
Z.~W. Brinbaum and O.~M. Klose.
\newblock Bounds for the variance of the mann-whitney statistic.
\newblock \emph{Annals of Mathematical Statistics}, 38, 1957.

\bibitem[Fonseca~et al(2017)]{PSGPPG}
P.~Fonseca~et al.
\newblock Validation of photoplethysmography-based sleep staging compared with
  polysomnography in healthy middle-aged adults.
\newblock \emph{Sleep}, 40, 2017.

\bibitem[Goldberger~et al(2000)]{Physionet}
A.~L. Goldberger~et al.
\newblock {PhysioBank, PhysioToolkit, and PhysioNet}: Components of a new
  research resource for complex physiologic signals.
\newblock \emph{Circulation}, 101:\penalty0 e215--e220, 2000.

\bibitem[Gotlibovych~et al(2018)]{Jawbone}
I~Gotlibovych~et al.
\newblock End-to-end deep learning from raw sensor data: Artial fibrillation
  detection using wearables.
\newblock \emph{KDD Deep Learning Day}, 2018.

\bibitem[Gupta \& Malik(2016)Gupta and Malik]{SupervisionTransfer}
Hoffman~J. Gupta, S. and J.~Malik.
\newblock Cross modal distillation for supervision transfer.
\newblock \emph{IEEE Conference on Computer Vision and Pattern Recognition},
  2016.

\bibitem[Hochreiter \& Schmidhuber(1997)Hochreiter and Schmidhuber]{LSTM}
S.~Hochreiter and J.~Schmidhuber.
\newblock Long short term memory.
\newblock \emph{Neural Computation}, 9(8):\penalty0 1735--1780, 1997.

\bibitem[Jan~et al(2019)]{EvalOfCoh}
H-Y Jan~et al.
\newblock Evaluation of coherence between {ECG} and {PPG} derived parameters on
  heart rate variability and respiration in healthy volunteers with/without
  controlled breathing.
\newblock \emph{Journal of Medical and Biological Engineering}, 39:\penalty0
  783--795, 2019.

\bibitem[Khandoker \& Palaniswami(2010)Khandoker and Palaniswami]{ApEvDiv}
Karmakar-C.~K. Khandoker, A.~H. and M.~Palaniswami.
\newblock Comparison of pulse rate variability with heart rate variability
  during obstructive sleep apnea.
\newblock \emph{Medical Engineering Physics}, 33(2):\penalty0 204--209, 2010.

\bibitem[LeCun~et al(1999)]{CNN}
Y.~LeCun~et al.
\newblock Object recognition with gradient-based learning.
\newblock \emph{Lecture Notes in Computer Science}, 1681, 1999.

\bibitem[Li~et al(2019)]{MicroSpeech1}
J.~Li~et al.
\newblock Large-scale domain adaptation via teacher-student learning.
\newblock \emph{Interspeech}, 2019.

\bibitem[Lin~et al(2014)]{ThatFrom}
W-H Lin~et al.
\newblock Comparison of heart rate variability from {PPG} with that from {ECG}.
\newblock \emph{The International Conference on Health Informatics}, pp.\
  213--215, 2014.

\bibitem[Long~et al(2018)]{CoDa}
M~Long~et al.
\newblock Conditional adversarial domain adaptation.
\newblock In \emph{Proceedings of the 32nd International Conference on Neural
  Information Processing Systems}, 2018.

\bibitem[Mandal \& Kent(2018)Mandal and Kent]{ApCorArt}
S.~Mandal and B.~D. Kent.
\newblock Obstructive sleep apnoea and coronary artery disease.
\newblock \emph{Journal of Thoracic Disease}, pp.\  S4212--S4220, 2018.

\bibitem[Meng~et al(2019)]{MicroSpeech2}
Z.~Meng~et al.
\newblock Domain adaptation via teacher-student learning for end-to-end speech
  recognition.
\newblock \emph{IEEE Automatic Speech Recognition and Understanding Workshop},
  2019.

\bibitem[Niculescu-Mizil \& Caruana(2005)Niculescu-Mizil and
  Caruana]{PlattScaling}
A.~Niculescu-Mizil and R.~Caruana.
\newblock Predicting good probabilities with supervised learning.
\newblock \emph{ICML}, 2005.

\bibitem[Pan \& Yang(2009)Pan and Yang]{SurvTranLer}
S.~J. Pan and Q.~Yang.
\newblock A survey on transfer learning.
\newblock \emph{IEEE Transactions on Knowledge and Data Engineering},
  22(10):\penalty0 1345--–1359, 2009.

\bibitem[Perez~et al(2019)]{AppleAfib}
M.~V. Perez~et al.
\newblock Large-scale assessment of a smartwatch to identify atrial
  fibrillation.
\newblock \emph{The New England Journal of Medicine}, 381:\penalty0 1909--1917,
  2019.

\bibitem[Punjabi(2008)]{OSAPrev}
N.~M. Punjabi.
\newblock The epidemiology of adult obstructive sleep apnea.
\newblock \emph{Proceedings of the American Thoracic Society}, pp.\  136--143,
  2008.

\bibitem[Purushotham~et al(2017)]{VarRecAdv}
S.~Purushotham~et al.
\newblock Variational recurrent adversarial deep domain adaptation.
\newblock In \emph{Proceedings of the International Conference on Learning
  Representations (ICLR)}, 2017.

\bibitem[Radha~et al(2018)]{PhilipsTL}
M~Radha~et al.
\newblock {LSTM} knowledge transfer for {HRV}-based sleep staging.
\newblock \emph{preprint, arXiv:1809.06221}, 2018.

\bibitem[Radha~et al(2019)]{Nature}
M.~Radha~et al.
\newblock Sleep stage classification from heart-rate variability using
  long-short-term-memory neural networks.
\newblock \emph{Nature Scientific Reports}, 9, 2019.

\bibitem[Saenko~et al(2010)]{AdaVizCats}
K.~Saenko~et al.
\newblock Adapting visual category models to new domains.
\newblock \emph{European conference on computer vision}, pp.\  213--226, 2010.

\bibitem[Scholkmann \& Wolf(2012)Scholkmann and Wolf]{AMPD}
Boss~J. Scholkmann, F. and M.~Wolf.
\newblock An efficient algorithm for automatic peak detection in noisy periodic
  and quasi-periodic signals.
\newblock \emph{Algorithms}, 5(4):\penalty0 588--603, 2012.

\bibitem[Tzeng~et al(2017)]{AdDiDoAd}
E~Tzeng~et al.
\newblock Adversarial discriminative domain adaptation.
\newblock In \emph{Proceedings of the International Conference on Learning
  Representations (ICLR)}, 2017.

\bibitem[Vandecasteele~et al(2017)]{Epilepsy}
K.~Vandecasteele~et al.
\newblock Automated epileptic seizure detection based on wearable {ECG} and
  {PPG} in a hospital environment.
\newblock \emph{Sensors}, 17, 2017.

\bibitem[Vaswani~et al(2017)]{AIAYN}
A.~Vaswani~et al.
\newblock Attention is all you need.
\newblock In \emph{Proceedings of the 31st International Conference on Neural
  Information Processing Systems}, 2017.

\bibitem[Wang \& Deng(2018)Wang and Deng]{DomainAdaptRev}
M.~Wang and W.~Deng.
\newblock Deep visual domain adaptation: A survey.
\newblock \emph{Neurocomputing}, 312:\penalty0 135--153, 2018.

\bibitem[Young~et al(1993)]{SDOB}
T~Young~et al.
\newblock The occurrence of sleep-disordered breathing among middle-aged
  adults.
\newblock \emph{New England Journal of Medicine}, pp.\  1230--5, 1993.

\end{thebibliography}
\bibliographystyle{iclr2020_conference}

\appendix
\section{Sleep apnea study details}
\label{section:apnea}
\subsection{Sleep Apnea}
Sleep apnea is a sleep disorder characterized by pauses in breathing or periods of shallow breathing during sleep. The most common form is obstructive sleep apnea (OSA), in which breathing is interrupted by a blockage of airflow. OSA is considered to be an underdiagnosed condition. Population prevalence is estimated at 9\%-50\%, and differs significantly between men and women and increases with age and obesity level \citep{OSAPrev}. OSA is highly correlated with stroke, hypertension, coronary artery disease and diabetes mellitus \citep{ApCorArt} (though the causal relationship is unclear). The gold standard diagnostic test for OSA in a clinical setting is a laboratory-based, overnights attended polysomnography (PSG) with expert annotation.
\subsection{Study description}

Data collection was carried out during a study of individuals presenting with symptoms of a sleep disorder, who were undergoing a diagnostic in-lab sleep study for clinical purposes.
Study subjects were assessed for sleep disorders via a single-night in-clinic polysomnography (PSG) test. After data quality filtering, a total of 17 subjects were included for analysis. The gender ratio of the subjects is 2:1 male to female which is consistent with what is reported in the literature \citep{SDOB}. The patient population was diverse with respect to age (between 18 and 75), ethnicity, and body mass index (BMI). All study participants wore a 
consumer-grade wrist-worn device with a 20Hz single wavelength photoplethysmography (PPG) sensor and a three-axis accelerometer. 

\subsection{Polysomnography} 
Ground truth apnea and hypopnea events were determined based on expert annotated flow events from PSG data collected on the same individuals which were recorded simultaneously to provide physiological measures of human activity and behavior during sleep. Scoring was completed according to AASM manual ver 2.4\citep{AASM} (1A rule) by a third party registered polysomnography technologist. The PSG sensors comprised of multiple electrical signals including electroencephalogram (EEG), electrooculogram (EOG), electromyogram (EMG) from both chin and legs, electrocardiogram (ECG), snore vibrations, breathing effort, blood oxygen saturation (SpO2) oral and nasal airflow.

\subsection{Data processing}
\label{subsection:preproc}
Signals from both the PSG device and the wrist-worn wearable were first roughly time-aligned using real-time clock (RTC) time stamps for the respective devices. However due to inaccuracies and drift in the RTCs, a finer synchronization was performed.

The general approach to data pre-processing is to extract time-domain beat/pulse wave fiducial markers from which we can obtain beat-to-beat or peak-to-peak heart rate intervals.

Peaks of each pulse wave recorded via the wrist-worn optical PPG sensor were detected using the Automatic Multi-scale Peak Detection (AMPD) algorithm \citep{AMPD}. The AMPD algorithm provides robustness to noise by automatically rejecting spurious peaks which are below an automatically selected scale. The algorithm works by constructing a local maxima scalogram which provides information about the frequency (count) of local maximas at various scales (separation) throughout the data. Once an optimal scale is found peak finding reduces to finding all maxima which persist up to that scale.

The largest amplitude feature in a typical electrocardiogram (ECG) lead is the QRS complex or simply R-wave which occurs during ventricular depolarization thus initiating contraction of the heart. We applied a simplified variant of the very common Pan-Tompkins algorithm which applies a band-pass filter, derivative filter, squaring operation and a moving window averaging to assess the signal (R-wave amplitude) vs noise level. The peak level associated with each peak is then compared to an adaptively updated threshold based on the noise level such that only unambiguous R-waves are detected using a fairly conservative SNR threshold. Given the general lack of gross body  motion and high quality DAC and recording properties of the PSG device the ECG is usually very high quality and low noise.


\section{PhysioNet data}
\label{subsection:pccc-data}
We used external data for domain adaptation from PhysioNet\citep{Physionet} which was contributed by the Massachusetts General Hospital’s (MGH) Computational Clinical Neurophysiology Laboratory (CCNL), and the Clinical Data Animation Laboratory (CDAC). The dataset includes 1,985 subjects which were monitored at an MGH sleep laboratory for the diagnosis of sleep disorders\footnote{\url{https://physionet.org/content/challenge-2018/1.0.0/}}. The arousal events were labelled by certified sleep technologists at the MGH based on the PSG waveforms and classified into spontaneous arousals, respiratory effort related arousals (RERA), bruxisms, hypoventilations, hypopneas, apneas (central, obstructive and mixed), vocalizations, snores, periodic leg movements, Cheyne-Stokes breathing or partial airway obstructions.

\label{section:synth}

\section{Ablation study}
\label{section:abl}
From the results of our main experiments (Table \ref{apnea-table}) it is clear that the basic aspects of the approach --- teacher supervision and transfer learning/domain adaptation --- are key for performance. Here we test the smaller details: we remove the output-prediction term from the loss function so that it matches the one in \citep{SupervisionTransfer}, remove the activation loss so that only the pre-trained models outputs are predicted, remove the CNN, and test a minimal implementation where the model's lower layers are simply fine-tuned to predict the pre-trained model's outputs. Results are shown in Table \ref{Ablation}. The minimal model performs worst and full model best. The CNN seems to give the worst effort-to-reward ratio, and can probably be removed if a simple implementation is desired. Means and standard deviations across five runs are reported: we see that the results are very stable.
\begin{table}[t]
\caption{Ablation study}
\label{Ablation}
\begin{center}
\begin{tabular}{lll}
\multicolumn{1}{c}{\bf MODEL}  &\multicolumn{1}{c}{\bf ROC AUC}  &\multicolumn{1}{c}{\bf PR AUC}
\\ \hline \\
Minimal  & 0.783 $\pm$ 0.006 & 0.432 $\pm$ 0.005 \\
No CNN & 0.791 $\pm$ 0.007 & 0.442 $\pm$ 0.006 \\
No Output Loss & 0.789 $\pm$ 0.005  & 0.448 $\pm$ 0.007 \\
No Activation Loss & 0.789 $\pm$ 0.005 & 0.448 $\pm$ 0.011 \\
Full Model & \bf{0.801} $\pm$ 0.007  & \bf{0.462} $\pm$ 0.006 \\
\end{tabular}
\end{center}
\end{table}

\section{Synthetic Data}
\begin{table}[t]
\caption{Per-window validation metrics for synthetic data}
\label{synth-table}
\begin{center}
\begin{tabular}{lll}
\multicolumn{1}{c}{\bf MODEL}  &\multicolumn{1}{c}{\bf ROC AUC}  &\multicolumn{1}{c}{\bf PR AUC}
\\ \hline \\
Label-Supervised (Simulated PPG Only) & 0.68 & 0.15 \\
Label-Supervised Transfer Learning & 0.71 & 0.20 \\
Label-Supervised Domain Adaptation & 0.73 & 0.20 \\
Teacher-Supervised (Simulated PPG Only) & 0.77 & 0.25 \\
Naive Transfer & 0.77  & 0.32\\
Teacher-Supervised Transfer Learning & \bf{0.81} & 0.37 \\
Teacher-Supervised Domain Adaptation  & \bf{0.81}  & \bf{0.38}\\
(ECG)             & 0.82 & 0.41 \\
\end{tabular}
\end{center}
\end{table}

To ensure that our method can be validated without access to proprietary data, we report the results of an easy-to-reproduce experiment on a synthetic dataset. We hold aside twenty subjects from PCCC, and simulate domain shift by randomly perturbing the ECG-derived peaks for these subjects. The perturbation is done according to a hidden Markov model, reflecting the bursts of noise and distortion found in the real data. The transition matrix we used is
\begin{equation}
\begin{bmatrix}
0.995 & 0.002 & 0.001 & 0.002 \\
0.01 & 0.98 & 0.01 & 0.0 \\
0.0 & 0.005 & 0.96 & 0.035 \\
0.03 & 0.0 & 0.97 & 0.0 \\
\end{bmatrix}.
\end{equation}
In state $i$, the probability of adding any noise at all is $p_i$. If noise is added, it is sampled from a normal distribution $\mathcal{N}(\mu_i, \sigma_i)$; the absolute value of the sample is taken before adding it to the signal (the absolute value prevents negative peak-to-peak intervals).
We used $p = (0, 0.5, 0.1, 0.7)$, $\mu = (\_, 0.2, 0.4, 0)$, $\sigma = (\_, 0.5, 0.6, 1)$.

Since there is no ``raw waveform" here, we omit the CNN: aside from that, we carry out the same experiments as in the sleep apnea study. The results, shown in Table \ref{synth-table}, are broadly similar to those from the real data. Although the labels have been (presumably carefully) prepared for a machine learning competition, the difference between label-supervised and teacher-supervised methods is even more pronounced here: the best label-supervised approach (domain adaptation) is worse than the worst teacher-supervised approach (straightforward supervised learning with teacher-supervision). The difference between domain adaptation and transfer learning (fine-tuning the entire model) is minimal on this data. This may be due to the omission of the CNN, which was shown to make a small positive difference in the ablation study on the real data (Table \ref{Ablation}). We also tried removing one or the other term from the loss function, which for this data made essentially no difference.

\end{document}